\documentclass[twocolumn]{article}

\usepackage{arxiv}
\usepackage[utf8]{inputenc} 
\usepackage[T1]{fontenc}    
\usepackage{hyperref}       
\usepackage{url}            
\usepackage{booktabs}       
\usepackage{tabularx}
\usepackage{caption}
\usepackage{amsfonts}       
\usepackage{nicefrac}       
\usepackage{microtype}      
\usepackage{lipsum}         
\usepackage{graphicx}
\usepackage{natbib}
\usepackage{doi}
\usepackage{balance}
\usepackage{amsmath,amsthm,amsfonts,amssymb,amscd}
\usepackage{multirow,booktabs}
\usepackage[table]{xcolor}
\usepackage{fullpage}
\usepackage{lastpage}
\usepackage{appendix}
\usepackage{enumitem}
\usepackage{fancyhdr}
\usepackage{mathrsfs}
\usepackage{wrapfig}
\usepackage{setspace}
\usepackage{calc}
\usepackage{multicol}
\usepackage{cancel}
\usepackage[retainorgcmds]{IEEEtrantools}
\usepackage{amsmath}
\usepackage{empheq}
\usepackage{framed}
\usepackage[most]{tcolorbox}
\usepackage{xcolor}
\usepackage[normalem]{ulem}
\usepackage{amsmath}
\usepackage{makecell}
\usepackage{listings}
\usepackage{stfloats}

\newcommand{\R}{\mathbb{R}}

\newcommand{\set}[1]{\{#1\}}

\renewcommand{\int}{\text{int}}

\title{Systematic Analysis of LLM Contributions to Planning: Solver, Verifier, Heuristic}

\renewcommand{\headeright}{}
\renewcommand{\undertitle}{}
\renewcommand{\shorttitle}{\textit{arXiv} Template}
\author{
    Haoming Li\textsuperscript{*}, Zhaoliang Chen\textsuperscript{*}, Songyuan Liu, Yiming Lu, Fei Liu \\
    Emory University, Department of Computer Science \\
    \texttt{hmlvenom@gmail.com, david.chen2@emory.edu, simon.liu@emory.edu,} \\
    \texttt{ylu456@emory.edu, fei.liu@emory.edu} \\
    \textsuperscript{*}These authors contributed equally to this work
}
\date{}

\begin{document}
\twocolumn[
\maketitle

\begin{abstract}
\noindent
In this work, we provide a systematic analysis of how large language models (LLMs) contribute to solving planning problems. In particular, we examine how LLMs perform when they are used as problem solver, solution verifier, and heuristic guidance to improve intermediate solutions. Our analysis reveals that although it is difficult for LLMs to generate correct plans out-of-the-box, LLMs are much better at providing feedback signals to intermediate/incomplete solutions in the form of comparative heuristic functions. This evaluation framework provides insights into how future work may design better LLM-based tree-search algorithms to solve diverse planning and reasoning problems. We also propose a novel benchmark to evaluate LLM's ability to learn user preferences on the fly, which has wide applications in practical settings. 

\end{abstract}

\vspace{1em}
]

\section{Introduction}
Large Language Models (LLMs) have revolutionized natural language processing, showcasing extraordinary capabilities in a wide range of tasks, from text generation to reasoning and knowledge retrieval. However, when it comes to solving problems that require structured planning or sequential decision-making, LLMs face inherent limitations. Their ability to generate coherent outputs often does not translate into systematic exploration or deliberate reasoning, which are essential for tasks involving complex problem-solving and multi-step planning.

To address these limitations, researchers have explored integrating LLMs with structured reasoning frameworks that provide the necessary scaffolding for planning. One promising avenue is the use of test-time scaling techniques, where additional computational resources are applied during inference to enable dynamic exploration and reasoning (\cite{snell2024scalingllmtesttimecompute}). These techniques often leverage tree-based search methods, such as Monte Carlo Tree Search (MCTS) or hierarchical frameworks, allowing LLMs to deliberate, evaluate, and backtrack over potential solutions in a structured manner (\cite{hao2023reasoninglanguagemodelplanning, yao2023treethoughtsdeliberateproblem, long2023largelanguagemodelguided}).

This integration has demonstrated remarkable success in enhancing LLM capabilities. By embedding LLMs within frameworks that emulate human-like trial-and-error reasoning, they can serve as heuristic functions or world models, guiding the search process and improving decision-making. Such approaches have been employed in various applications, ranging from mathematical problem-solving to creative tasks, highlighting their potential to expand the scope of LLMs in domains where raw generative abilities are insufficient.

In this work, we present the following key contributions:
\begin{itemize}[leftmargin=*]
    \item We establish a systematic framework to evaluate the contributions of LLMs when integrated with structured algorithms
    \item We empirically validate the widely held assumption that LLMs can serve effectively as heuristic functions within tree-based algorithms. While this assumption underpins many existing methods, it is not a natural consequence of LLM design and requires rigorous testing to confirm.
    \item We propose a novel benchmark for evaluating LLMs' ability to learn user preferences. This benchmark has significant practical implications, as preference modeling is crucial in applications ranging from personalized recommendations to decision-support systems.
\end{itemize}

By addressing these dimensions, our work advances the understanding of how LLMs can be systematically integrated into planning and decision-making frameworks, paving the way for more robust and adaptable AI systems.

\section{Related Works}

\subsection{LLM in Planning}
There has been a lot of work focusing on the planning abilities of LLMs and designing systems that use LLMs as a part of core mechanism. A key advantage of LLMs is their capacity to interpret and generate natural language, enabling them to craft plans without extensive domain-specific training. However, their inherent non-determinism introduces challenges in predictability, unlike symbolic planners. Techniques like Reflexion \cite{shinn2023reflexion} and ReAct \cite{yao2023react} leverage iterative feedback loops and interleaved reasoning-action frameworks, respectively, to refine plans dynamically, reducing errors like hallucinations. Similarly, methods like Translated $\langle LM\rangle$ \cite{huang2022language} and LLM-Planner \cite{song2023llmplanner} enhance planning by generating free-form action plans and adjusting based on real-world observations or semantic translations. 

For structured task decomposition, approaches like SELFGOAL \cite{yang2024selfgoallanguageagentsknow} decompose complex goals into actionable subgoals using modules powered by LLMs, balancing decomposition with post-hoc summarization to guide agent behavior in multi-agent scenarios. Tree of Thoughts \cite{yao2023treethoughtsdeliberateproblem} and RAP \cite{hao2023reasoninglanguagemodelplanning} extend LLM reasoning through tree-based search strategies, with the latter incorporating Monte Carlo Tree Search for strategic exploration. Heuristic approaches, such as SayCanPay \cite{hazra2024saycanpay}, further blend LLM-generated actions with feasibility assessments to achieve grounded and cost-effective plans.

Additionally, code-based planning systems such as PROGPROMPT \cite{singh2022progprompt} and AdaPlanner \cite{sun2023adaplanneradaptiveplanningfeedback} employ programming-inspired prompts and adaptive feedback loops to generate executable plans that respond dynamically to environmental feedback. Separately trained components like those in LID \cite{li2022lid} and DEPS \cite{wang2023describe} combine pre-trained language models with multimodal systems or selector modules, enabling iterative plan refinement through active data gathering or error explanations. Collectively, these diverse strategies illustrate the transformative potential of LLMs in adaptive, interactive planning while addressing challenges in determinism and execution fidelity.

\subsection{Test-time Scaling}
The key concept behind test-time scaling is intuitive -- solving harder problems should not require the same amount of compute as solving easier ones. Most of the approaches described in the previous section already employs test-time scaling. Indeed, methods like ReAct \cite{yao2023react}, Reflexion \cite{shinn2023reflexion}, ToT \cite{yao2023treethoughtsdeliberateproblem} uses multiple interactions with LLM to get a final solution. In general, any algorithms that involve tree search, reflection mechanisms, or iterative feedback inherently embraces the concept of test-time scaling. More recently, OpenAI's new model o1 has demonstrated remarkable capabilities \cite{openai2024learning}. While the exact technical details are unpublished, it is evident that o1 leverages more test-time compute to generate bette solutions. 

There has been an increasing interest in systematically understanding this technique. A recent paper from DeepMind \cite{snell2024scalingllmtesttimecompute} emphasizes that allocating additional computational resources during inference allows models to refine their outputs iteratively, leading to substantial performance gains in tasks requiring complex reasoning. This "compute-optimal" strategy highlights a cost-effective approach that prioritizes flexible deployment over expensive pre-training cycles. By focusing on how models process and reprocess input during inference, this line of research shifts attention from scaling the number of parameters to scaling computational intensity during test-time, demonstrating that efficient problem-solving is achievable even with existing architectures. Similarly, \textit{Large Language Monkeys}\cite{brown2024largelanguagemonkeysscaling} underscores the impact of generating and sampling multiple outputs, showing a log-linear improvement in problem coverage with increased inference sampling. Together, these studies establish test-time computation as a powerful lever for enhancing LLM accuracy and adaptability. They also underline the challenges of balancing computational overhead with result quality, especially in tasks without automated solution verification. This burgeoning field paves the way for more efficient and versatile LLM deployments across diverse applications.
\balance 

\section{Evaluation Method}

We now introduce the proposed evaluation framework. We specifically focus on 3 different roles of LLM that may contribute to solving a planning task -- solver, verifier, and comparative heuristic function. We emphasize that these are independent components that can be composed together in very interesting ways. It is also possible to introduce external tools, such as symbolic planners, into the planning algorithm to further enhance performance and efficiency. 

We test those independent components on 3 tasks: Travel Planning, Course Planning, and Fitness Planning. We use TravelPlanner \cite{xie2024travelplanner} for the first task, and construct our own datasets for the other two. Notably, Fitness Planning benchmark is an interactive environment that tests LLM's ability to learn user preferences, instead of a fixed set of problems. 

On these 3 datasets, we have natural ways to define how "close" a proposed solution is to the actual solution for any given problem instance. This means that we can define an oracle heuristic function for each dataset, which is used to evaluate how well LLM-parameterized heuristic performs. 

\subsection{LLM as Solvers}
The most natural way to use LLMs in a planning problem is to simply ask the LLM to solve it. Therefore, as the first part of our evaluation, we test how well does LLM solve the problem. In our experiments, we use both direct prompting and CoT. The metric depends on the specific dataset.

\subsection{LLM as Verifiers}
We also explore whether LLMs can reliably verify the solution of a problem. Earlier work such as LLM Monkeys \cite{brown2024largelanguagemonkeysscaling} have shown that given sufficiently many attempts, LLMs are generally able to get the correct answer. Identifying the correct solution out of all generated responses without an automatic evaluation engine is essential, since in many real world applications, such oracle verifier is usually not available. 

A natural option to choose a solution out of $N$ is through majority voting, but as \cite{brown2024largelanguagemonkeysscaling} demonstrates, this technique is not powerful enough and does not scale very well with $N$. An intuitive explanation is that this method only works if the correct solution has high likelihood to begin with, which is not necessarily the case, especially for hard problems. In this paper, we evaluate whether LLMs can serve as verifiers out of the box. 


\subsection{LLM as Heuristic Functions}


\textbf{Real-valued Formulation.} For task $T$, we define an oracle heuristic function $f_T(\hat{y}, y, x) \to \R$, where $y$ is a ground truth solution, $\hat{y}$ is LLM-generated solution, and $x$ is problem description. We want to evaluate LLM's performance as an approximate of $f_T$, denoted $\hat{f}_T(\hat{y}, x) \to \R$, whose output does not rely on ground truth solution $y$. We can use $|f_T(\hat{y}, y, x) - \hat{f}_T(\hat{y}, x)|$ as a metric to evaluate LLM's performance. 

\textbf{Issues with real-valued formulation.} Estimating the exact value produced by $f_T$ may prove difficult. The range of such value is also highly dependent on the problem instance. For example, if we use editing distance as $f_T$ for course planning, then the output range is strictly integers within $(0, \text{total number of classes})$. This example also sheds light on the difficulty of estimating $f_T$. Indeed, without ground truth solution $y$, solving for editing distance could be as complex as solving the original problem directly, making this estimated heuristic unuseful. In short, in a real-valued formulation, we require $f_T$ to be 1) powerful enough to provide useful guidance for planning and 2) sufficiently independent of solution $y$ such that LLM can reasonably estimate $f_T$ without $y$. Finding such $f_T$ could be very difficult.

\textbf{Comparative Formulation}. One of the main purposes of having a heuristic function is to guide LLM's planning/reasoning process through tree search, where the solver LLM proposes $N$ candidate next steps and heuristic LLM $\hat{f}(\hat{y}, x)$ evaluates each candidate. However, observe that the exact values are not strictly necessary here to provide guidance (although having it could be beneficial). Indeed, to effectively search the tree, providing an \textbf{ordering} of candidate solutions is sufficient. Let $T$ be a task, we define a (ground truth) comparison function $g_T(\hat{y}_1, \hat{y}_2, y, x) \to \set{0, 1}$, where $\hat{y}_1, \hat{y}_2$ are candidate (intermediate) solutions, $y$ is the ground truth solution, and $x$ is problem description. A simple implementation of such function is
\begin{equation}
\begin{split}
        g_T(\hat{y}_1, \hat{y}_2, y, x) &= \text{argmax}([f_T(\hat{y}_1, y, x), f_T(\hat{y}_2, y, x)])\\
    &= \begin{cases}
    0 & \text{ if } f_T(\hat{y}_1, y, x) \ge f_T(\hat{y}_2, y, x)\\
    1 & \text{ otherwise }
\end{cases}
\end{split}
\end{equation}
With pair-ordering function well defined, we can easily extend it to comparing $N$ candidates. Our goal is to evaluate how close LLM's approximation, denoted $\hat{g}_T(\hat{y}_1, \hat{y}_2, x)$, is to $g_T$. Note that as before, $\hat{g}_T$ does not rely on $y$. The subtlety here is that because we only care about ordering and not exact value, it could be much easier for LLM to do. Because LLM no longer needs to explicitly predict the value of $f_T$, we alleviate the restrictions posed on $f_T$ that was detailed earlier. Using the running example of course planning problem, of two feasible solutions, the one with higher preference score may be more optimal than the other. As another example, LLM may observe that one of the candidate intermediate solutions is already infeasible because there is an unplanned course that cannot fit into the schedule due to time constraints. Knowing that this candidate should be ordered lower than the other is much simpler than predicting the editing distance to make it right.

\section{Datasets}

\subsection{Fitness Planning}

Previous studies focusing on personal fitness planning such as PlanFitting\cite{shin2023planfittingtailoringpersonalizedexercise} provides a comprehensive framework for implementing planning on user's end. However, evaluation of LLM performance on such a system depends on continuous human feedback, which is both costly and time-inefficient. In this paper, we propose a new framework for evaluating LLM performance on personal fitness planing.

The goal of this problem is to optimize a workout plan for a user by adjusting assigned exercises and corresponding repetitions (reps) iteratively, based on the feedback provided by the user after each workout session. The agent is tasked with finding the optimal plan by exploring various exercise combinations under a set of constraints, while continuously refining the plan based on user satisfaction. Specifically, within the task, there will be two entities, a \textbf{User} who provides basic information and feedback (satisfaction scores, complaints, etc.) after completing workout plans, and an \textbf{Agent} who generates initial and updated workout plans, aiming to maximize user satisfaction through iterative feedback.

To modularize the problem for flexible adjustments, we divide the problem into four major components: the user bank, exercise bank, emergency condition bank, and evaluation method, all of which will bring variations to the plan optimization iterations.

\textbf{Problem Formulation}  Our objective is to maximize \textbf{User} satisfaction, denoted as \( f(P_t, \hat{P_t}) \rightarrow F_t \), subject to constraints \( C(B, Z) \), where $f$ is hidden from both the \textbf{agent} and the \textbf{user}. Specifically:

\begin{itemize}[leftmargin=*]
    \item \( P_t \): The fitness plan at timestep \( t \), represented as \( P_t = \langle p_{1,t}, p_{2,t}, \dots, p_{k,t} \rangle \), where \( k \) is the number of exercises, and \( p \) refers to the reps assigned to each exercise.
    \item \( U \): The user preference, represented as \( U = \langle u_1, u_2, \dots, u_k \rangle \), $\forall u \in [0, 10]$ where \( k \) is the number of exercises, and \( u \) refers to the \textbf{user}'s preference to each exercise.
    \item $\hat{P_t}$: The desired fitness plan at time step $t$ given the generated plan $P_t$, represented as \( \hat{P_t} = \langle \hat{p}_{1,t}, \hat{p}_{2,t}, \dots, \hat{p}_{k,t} \rangle \). It can also be considered as the target the agent is trying to uncover. The specific values of $\hat{p}_{i,t}$ is calculated by optimizing the below expression:
    \begin{equation}
        \begin{split}
            \text{Maximize } &\sum_{i} u_i \cdot \hat{p}_{i,t}\\
            \text{Subject to } &\sum_{i} t_i \cdot p_i \leq T, \\
            &0 \leq \hat{p}_{i,t} \leq max\_reps
        \end{split}
    \end{equation}
    where $t_i$ is the time required to complete exercise $p_i$, $T$ is the total available time of the user.
    \item \( F_t \): The user feedback at timestep \( t \). This can either be a real number in the range [0, 10], representing the overall satisfaction with \( P_t \), or a set of real numbers in the same range, reflecting feedback on specific attributes of \( P_t \) (e.g., satisfaction with individual exercises).
    \item \( B \): A set of boolean constraints on \( P_t \), such as the availability of gym equipment.
    \item \( Z \): A set of integer constraints on \( P_t \), such as the \textbf{user}'s available time or stamina.
    \item \( f \): The utility function that maps $P_t, \hat{P_t}$ to a real number, or the satisfaction score. It is consisted of three components, evaluation on selected exercise, assigned reps, and diversity. We argue that such composition suits with real-world workout planning. 
    \[
        \text{Plan}(P_t, U) = \frac{1}{10}\frac{\sum_{i=1}^n U_i \cdot \mathbb{I}(P_{t,i} \neq 0)}{\sum_{i=1}^n \mathbb{I}(P_{t,i} \neq 0)}
    \]
    
    \[
        \text{Rep}(P_t, \hat{P_t}) = \left( \frac{P_t \cdot \hat{P_t}}{\|P_t\| \|\hat{P_t}\|} \right) \cdot \min\left(1, \frac{\|P_t\|}{\|\hat{P_t}\|}, \frac{\|\hat{P_t}\|}{\|P_t\|}\right)
    \]
    \[
        \text{OverLap}(P_{t-k}, \dots, P_t) = -\frac{1}{n}\sum_{i=1}^n \mathbb{I}\left(\bigcup_{j=0}^k P_{t-j,i} \neq 0\right)
    \]
    Since the output of all three functions are with the range [0, 1], the final score will be the weighted sum of the three evaluation functions multiplying 10 (max satisfaction score), controlled by two hyperparameters $\alpha$ and $\beta$.
    \begin{equation}
        \begin{aligned}
        F_t = 10 \cdot \Big(
            & \alpha \cdot \text{Plan} + \beta \cdot \text{Rep} \\
            & + (1 - \alpha - \beta) \cdot \text{Overlap}
        \Big)
        \end{aligned}
    \end{equation}

\end{itemize}

\begin{figure*}[htbp!]
\centering
\includegraphics[width=\linewidth]{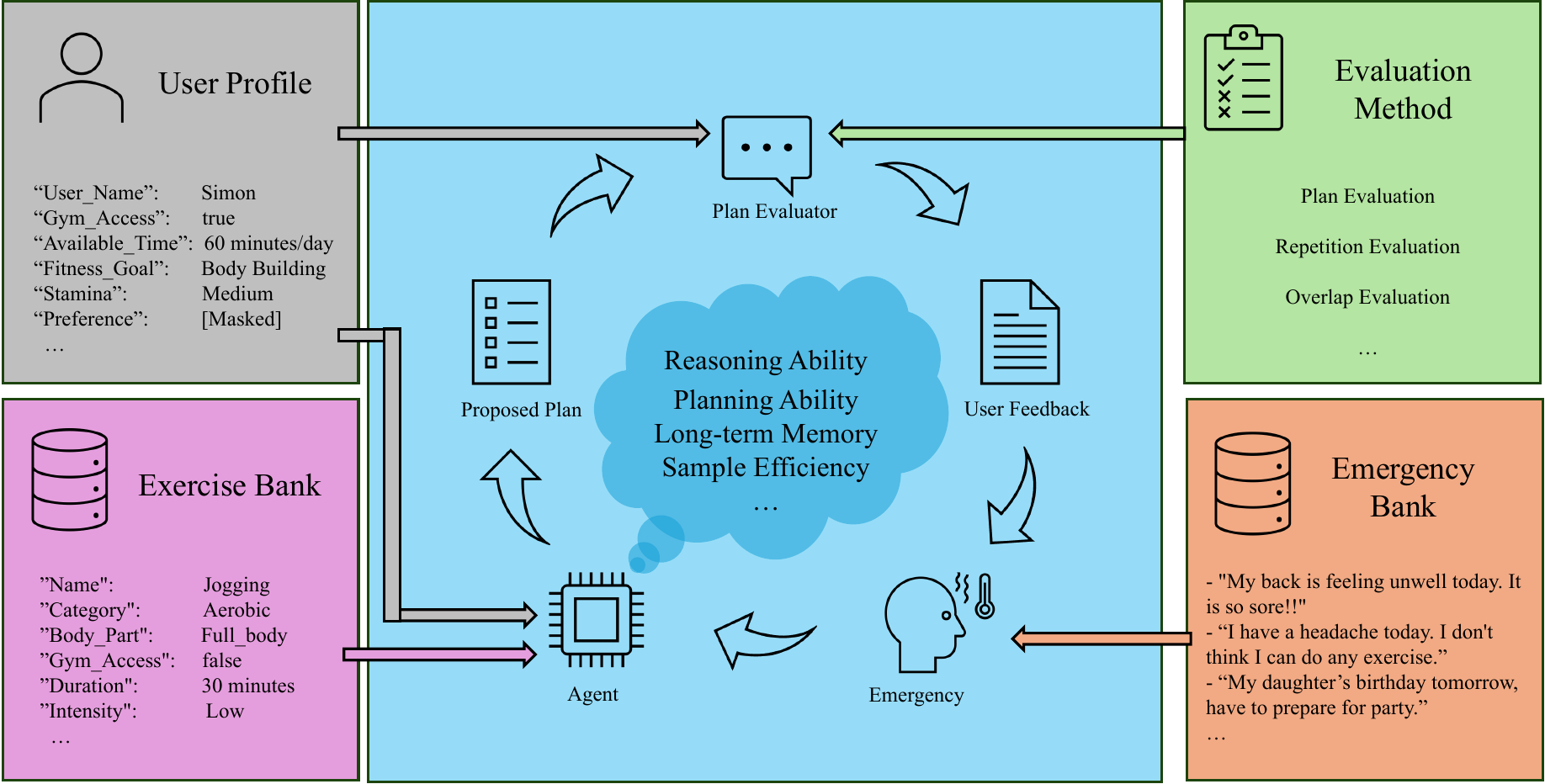}
\vspace{-0.1cm}
\caption{Illustration of Proposed Fitness Planning Framework}
\label{fig:fitness_illu}
\vspace{-0.3cm}
\end{figure*}

At each iteration \( t \), the \textbf{agent} utilizes its reasoning and planning capabilities to learn a "plan generating function" \( g(\{P_1, \dots, P_{t-1}\}, \{F_1, \dots, F_{t-1}\}) \rightarrow P_t \), where \( P_t \) is constrained by \( C(B, Z) \). The agent outputs a fitness plan \( P_t = \langle p_{1,t}, p_{2,t}, \dots \rangle \) designed to maximize the \textbf{user} utility function \( f \). The agent aims to balance the feedback from previous iterations to explore different exercises and converge on an optimal plan. An example of the our proposed fitness planning is shown in Appendix A.

Previous studies have showed that LLMs suffer from providing remedies to plans using newly introduced emergency events, such as illness.\cite{shin2023planfittingtailoringpersonalizedexercise} Therefore, we take such emergency scenarios into consideration by introducing dynamic constraints. During the iterative process, dynamic constraints may be introduced as part of user feedback. For example, if the user experiences back pain due to poor sleep, the agent must avoid exercises that engage the back muscles.

Let \( C_d \) represent such dynamic constraints that will be introduced under probability $p$. The agent must adapt the plan accordingly, incorporating both boolean and real-value constraints as new information becomes available. Specifically, the plan generated at next time-step $P_t$ will be subject to constraints $\hat{B}$ and $\hat{C}$, where $\hat{B} = B \cup C_d$, $\hat{Z} = Z \cup C_d$ depending on the attribute of $C_d$.

\textbf{Evaluating Solver} The optimization process will be measured by the following metrics: 1) \textit{Feasibility}: Percentage of time the {Agent} generates a feasible plan; 2) \textit{Optimality}: How closely the plan matches with the user's preferences in the last three days; 3) \textit{Cost Utility}: How many iterations does the {Agent} take to reach an optimal solution; and 4) \textit{Diversity}: The diversity of exercises explored by the {Agent} during the iterations.

\textbf{Evaluating Verifier} We define verifier for fitness planning under two settings, zero-shot and few-shot verification. In zero-shot setting, the agent will be provided with the initial plan $P_0$, the user preference $U$, the constraints $B$ and $Z$. It will then be asked to verify if $P_0$ is admissible. At few-shot setting, the agent will be additionally provided with the $P_{1}, ..., P_{t}$ and $F_{0}, ..., F_{t-1}$. Alternatively, the agent will be asked to verify the admissibility of the plan $P_t$. We evaluate the performance of LLMs using passrate, which measures whether a plan satisfies all given constraints.

\textbf{Evaluating LLM Heuristic} Similar to the verifier, we define the heuristic function for fitness planning under two settings, zero-shot and few-shot heuristic ranking. In zero-shot setting, the agent will be provided with $n$ initial plans $P_0^0, ..., P_0^n$, the user preference $U$, the constraints $B$ and $Z$. It will then be asked to rank the plans under our heuristic definition. At few-shot setting, the agent will be additionally provided with the $P_{1}, ..., P_{t}$ and $F_{0}, ..., F_{t-1}$. Alternatively, the agent will be asked to rank the plans $P_t^0, ..., P_t^n$.

\subsection{Course Planning}
Although previous paper \cite{kasneci2023chatgpt} has emphasized the opportunities and challenges of applying LLMs for lesson planning, which is to generate a syllabus for a particular course, few words mentioned the application of combining LLMs with course schedule planning. In this paper, we propose a new benchmark to address this blank.

The goal of this problem is to test LLMs' ability to generate and evaluate a plan that arranges different sections of courses into a calendar. Specifically, there are several available classrooms $(C_1,C_2,...,C_n)$, each with a specific seating capacity $(c_1,c_2,...,c_n)$, and multiple course sections$(S_1,S_2,...,S_m)$, each assigned a specific time slot and set of days for their scheduled classes, with each section also having an estimated enrollment $(s_1,s_2,...,s_m)$, waiting to be assigned to appropriate classrooms. The agent is going to assign classrooms to each section and to get the overall classroom occupancy above a certain threshold.

\textbf{Problem Formulation} The objective of this task is to generate a plan $P=(p_1,p_2,...,p_n)$ where $p_i=(s_{i,j},c_{i,k})$ is a section-classroom pair chosen from every action $i$ and try to get as high a classroom occupancy as possible so that the overall occupancy exceeds a certain threshold value. 
\[
J(P) = \sum_{i=0}^nc_{i,k}-s_{i,j} > \delta
\]

Specifically, $s_{i,j}$ represent the estimated enrolled number of students in the $j_{th}$ section during action $i$, and $c_{i,k}$ denote the seating capacity of the $k_{th}$ classroom in action $i$. The plan should adhere to the following constraints:

\begin{enumerate}[leftmargin=*]
    \item \textbf{No conflicts in classroom usage at the same time}:
    \[
    \sum_j x_{k,j,t} \leq 1, \quad \forall k, \forall t
    \]
    where $x_{k,j,t} = 1$ if classroom $k$ is assigned to section $j$ at time $t$, and $0$ otherwise.
    
    \item \textbf{Classroom capacity is sufficient for assigned sections}:
    \[
    s_{i,j} \leq c_{i,k}, \quad \forall i, \forall j, \forall k
    \]

    \item \textbf{Each section must be assigned to exactly one classroom}:
    \[
    \sum_k z_{j,k} = 1, \quad \forall j
    \]
    where $z_{j,k} = 1$ if section $j$ is assigned to classroom $k$, and $0$ otherwise.
\end{enumerate}

\textbf{Evaluating Solver}
As the problem is NP-hard, it's hard to find all possible answers in polynomial time, therefore, we measure the generated plan to be an answer by three metrics: 1) \textit{Completeness}: if the generated plan contains all sections.(fulfill constraint 3) 2) \textit{Feasibility}: for those completed plans, whether there is no time conflict and whether the classroom capacity is no smaller than estimated enrollment.(fulfill constraint 1 and 2) 3)\textit{Optimality}: for those feasible plans, whether the overall classroom occupancy exceeds the threshold $\delta=1.3$. 

\textbf{Evaluating Verifier}
To evaluate the ability of LLMs to verify whether a given plan is a valid solution, the LLMs are tasked with assessing candidate plans, which are either generated by other LLMs or derived from the ground truth pool through exhaustive search to obtain a valid solution. We evaluate LLM as verifier through two metrics: \textit{feasibility}: whether the assessment of the feasibility of the given plan is accurate., \textit{Optimality}: for feasible plans, whether the assessment of their optimality is accurate.

\textbf{Evaluating LLM using Heuristic}
To evaluate the ability of LLMs using heuristic, we define the task under two settings, zero-shot and one-shot. In either setting, LLMs are tasked to give a rank over two or four candidate plans. These candidate plans ($\hat{P}_{1}, \hat{P}_{2},...,\hat{P}_{n}$) are derived from a correct plan $P_{gold}$, each representing an intermediate state in the process of arriving at the correct solution. Additionally, the content of these plans is randomly altered at a certain rate to produce $\hat{P}_{i}$. The goal of this task is to test whether LLMs can detect the incompleteness and incorrectness of a plan and estimate a distance $dist(\hat{P}_i, P_{gold})$ from current plan to the correct plan. In the one-shot setting, an example containing two candidate plans and a comparative heuristic function is given. LLMs can follow the idea and thought of the heuristic to determine the rank of plans. We measure the performance of LLM through $hit@k$ metrics, which evaluates whether at least one correct answer is included in the top $k$ answers given by LLMs.

\textbf{Dataset}
Based on the number of courses, the dataset is separated into three levels of difficulty, with each level we have 400 instances. Each course contains two or three sections, each of which contains 20 to 30 students. Each level contains three to six classrooms with each classroom containing 25 to 35 seats and while the difficulty goes up, the number of sections has increased dramatically compared to the number of classrooms, making it harder to plan and assign classrooms to sections. Table \ref{tab:describe} shows basic information of the dataset.

\begin{table}[h!]
\centering
\caption{Course and Classroom Statistics}
\renewcommand{\arraystretch}{1.3} 
\begin{tabularx}{0.5\textwidth}{@{}lccc@{}}
\toprule
\textbf{Indicator} & \textbf{Easy} & \textbf{Medium} & \textbf{Hard} \\ 
\midrule
\# Courses              & 5        & 7        & 10       \\ 
Avg \# Sections         & 2.50     & 2.49     & 2.48     \\ 
Avg \# Students/Sec     & 24.85    & 24.91    & 24.95    \\ 
Avg \# Classrooms       & 4.03     & 4.53     & 4.97     \\ 
Avg \# Seats/Classroom  & 29.73    & 29.85    & 29.78    \\ 
Sections / Classroom    & 3.39     & 4.03     & 5.02     \\ 
\bottomrule
\end{tabularx}
\label{tab:describe}
\end{table}

\subsection{Travel Planning}
The goal of TravelPlanner dataset is to test LLM's ability to generate comprehensive travel itineraries while satisfying user-specified needs and commonsense constraints. Each task is initiated by a user query that provides details such as departure and destination cities, travel dates, budget, number of travelers, and specific preferences like accommodation or transportation needs. The agent gathers the necessary information using a set of tools, such as a flight search tool for retrieving flight details between cities, and must create a travel plan that meets all requirements. An easier version of the problem allows the agent to focus solely on planning by directly providing the information that would otherwise be retrieved via tools, removing the need for active data collection.

The performance of the agents is evaluated across several key metrics: \textit{Delivery Rate} (whether the agent can produce a plan within a fixed number of steps), \textit{Hard Constraint Pass Rate}, \textit{Commonsense Constraint Pass Rate}, and \textit{Final Pass Rate} (percentage of plans that pass both hard and commonsense constraint). The \textit{hard constraints} are user-specified and include staying within the specified budget, matching the user's accommodation preferences (e.g., type of room or pet-friendliness), etc. In addition, the problem involves a set of \textit{commonsense constraints}, such as ensuring that the plan contains all necessary information (e.g., flights and accommodations), that the order of city visits follows a logical route, that transportation modes are non-conflicting, and that the selected restaurants and attractions are not repeated during the trip.

\textbf{Evaluating LLM Heuristic} A simple oracle heuristic function is to compute the micro pass rate of all constraints in the plan generated so far. This can be done by adjusting the automated plan evaluation script provided by the authors. A higher pass rate means that a proposed plan is closer to the ground truth solution. Given 2 plans, the one with higher pass rate is the better plan.

\section{Experimental Results}

In this section, we present main experimental findings from our proposed evaluation framework. We use claude-3-5-sonnet-20240620, GPT-4o-2024-08-06, and DeepSeek-V2.5 to show that the same pattern manifests across different models. The detailed report for LLMs work as solver can be find in \ref{tab:solver_datasets}.

\begin{table}[!htbp]
    \centering
    \begin{tabular}{l ll l r}
    \toprule
    \textbf{Dataset} & \multicolumn{2}{c}{\textbf{Setup}} & \textbf{Model} & \textbf{Passrate}\\
    \midrule
    
    \multirow{9}{*}{Course} 
      & \multicolumn{2}{c}{}   & GPT-4o    & 0.0250 \\
      & \multicolumn{2}{c}{Easy}       & Claude    & 0.0725 \\
      & \multicolumn{2}{c}{}       & DeepSeek  & 0.0125 \\
      \cmidrule(lr){2-5}
      & \multicolumn{2}{c}{} & GPT-4o    & 0.0075 \\
      & \multicolumn{2}{c}{Medium}       & Claude    & 0.0275 \\
      & \multicolumn{2}{c}{}       & DeepSeek  & 0.0025 \\
      \cmidrule(lr){2-5}
      & \multicolumn{2}{c}{}   & GPT-4o    & 0.0000 \\
      & \multicolumn{2}{c}{Hard}       & Claude    & 0.0025 \\
      & \multicolumn{2}{c}{}       & DeepSeek  & 0.0025 \\
    
    \midrule
    \multirow{12}{*}{Fitness} 
      & \multirow{6}{*}{without} & \multirow{2}{*}{5 iter}  & GPT-4o   & 0.7660 \\
      &                          &                           & DeepSeek & 0.0000 \\
      \cmidrule(lr){3-5}
      &                          & \multirow{2}{*}{10 iter} & GPT-4o   & 0.8490 \\
      &                          &                           & DeepSeek & 0.0000 \\
      \cmidrule(lr){3-5}
      &                          & \multirow{2}{*}{20 iter} & GPT-4o   & 0.8520 \\
      &                          &                           & DeepSeek & 0.0000 \\
      
      \cmidrule(lr){2-5}
      
      & \multirow{6}{*}{with}    & \multirow{2}{*}{5 iter}  & GPT-4o   & 0.5000 \\
      &                          &                           & DeepSeek & 0.0000 \\
      \cmidrule(lr){3-5}
      &                          & \multirow{2}{*}{10 iter} & GPT-4o   & 0.6140 \\
      &                          &                           & DeepSeek & 0.0000 \\
      \cmidrule(lr){3-5}
      &                          & \multirow{2}{*}{20 iter} & GPT-4o   & 0.7050 \\
      &                          &                           & DeepSeek & 0.0000 \\
    
    \midrule
    \multirow{4}{*}{Travel} 
      & \multicolumn{2}{c}{Direct}  & GPT-4o    & 0.0720 \\
      & \multicolumn{2}{c}{}        & DeepSeek  & 0.0050 \\
      \cmidrule(lr){2-5}
      & \multicolumn{2}{c}{CoT}     & GPT-4o    & 0.0830 \\
      & \multicolumn{2}{c}{}        & DeepSeek  & 0.0000 \\
      
    \bottomrule
    \end{tabular}
    \label{tab: simple_solver}
    \vspace{0.5em}
    \caption{LLMs Work as Solver Across Multiple Datasets}
\end{table}

\subsection{Fitness Planning}

In Table \ref{tab:solver_datasets}, GPT-4o demonstrates superior performance compared to DeepSeek across all conditions. Without dynamic constraints, GPT-4o achieves high feasibility, optimality, and diversity scores. However, an interesting observation arises at 10 iterations, where the feasibility score (0.6670) is slightly lower than at 5 iterations (0.7370). This anomaly suggests that while additional iterations allow for more exploration and refinement of plans, they may occasionally lead to a trade-off in feasibility due to overfitting or over-exploration of the solution space. By 20 iterations, GPT-4o recovers, achieving its highest feasibility (0.7500) and optimality (0.8520) in this condition, showing that longer iterations generally benefit the model's performance.

Under dynamic constraints, both models experience performance degradation, with GPT-4o showing significant drops in feasibility and optimality. At 5 iterations, feasibility is reduced to 0.4130, but it gradually improves to 0.6300 by 20 iterations. This recovery demonstrates GPT-4o's ability to adapt to stricter constraints with more time. However, diversity scores remain lower compared to the unconstrained setting, reflecting the challenge of balancing exploration with adherence to dynamic requirements. In contrast, DeepSeek fails to produce any feasible outputs under either condition, indicating its limited capability to handle both unconstrained and constrained tasks.

In Table \ref{tab:verifier_datasets}, both LLMs demonstrate strong performance with high pass rates, highlighting their ability to verify fitness plans effectively under the given settings. Notably, the performance on fitness planning surpasses other two datasets. This improved performance can be attributed to the relatively simpler structure and fewer constraints in the fitness planning task which involves fewer overlapping constraints and a more straightforward evaluation of user preferences, allowing both GPT-4o and DeepSeek to achieve higher pass rates with less computational overhead.

Table \ref{tab:heuristic_4candidate} illustrates that both DeepSeek and GPT-4o display competitive abilities in identifying top plans. DeepSeek maintains a slight edge in pinpointing the optimal plan, but GPT-4o’s higher comparison accuracy suggests it offers more nuanced discernment when evaluating similar plans. This subtle advantage indicates that although GPT-4o may not always lead in initial correctness, its refined comparative judgment allows it to more effectively distinguish closely matched fitness solutions and thereby provide a more balanced evaluative performance.

\begin{table*}[t]
\centering
\caption{LLMs Work as Verifier Across Multiple Datasets}
\renewcommand{\arraystretch}{1.3} 
\begin{tabularx}{\textwidth}{@{}l>{\centering\arraybackslash}Xccc@{}}
\toprule
\textbf{Dataset} & \textbf{Model} & \textbf{Feasibility} & \textbf{Optimality} & \textbf{Pass Rate} \\ 
\midrule
\multirow{3}{*}{Course Planning-Easy} 
    & GPT-4o    & 0.5825 & 0.1374 & 0.3075      \\ 
    & Claude    & 0.5725 & 0.9336 & 0.3200      \\ 
    & DeepSeek  & 0.5675 & 0.0569 & 0.2950      \\ 
\midrule
\multirow{3}{*}{Course Planning-Medium} 
    & GPT-4o    & 0.5500 & 0.0583 & 0.3000      \\ 
    & Claude    & 0.5450 & 0.9757 & 0.2800      \\ 
    & DeepSeek  & 0.5475 & 0.0340 & 0.2925      \\ 
\midrule
\multirow{3}{*}{Course Planning-Hard} 
    & GPT-4o    & 0.5150 & 0.0308 & 0.2700      \\ 
    & Claude    & 0.5550 & 0.9744 & 0.3025      \\ 
    & DeepSeek  & 0.5425 & 0.0103 & 0.2950      \\ 
\midrule
\multirow{2}{*}{Fitness Planning} 
    & GPT-4o    & -      & -      & 0.8400   \\ 
    & DeepSeek  & -      & -      & 0.8200   \\ 
\midrule
\multirow{2}{*}{Travel Planning} 
    & GPT-4o    & -      & -      & 0.4550 \\ 
    & DeepSeek  & -      & -      & 0.4890 \\ 
\bottomrule
\end{tabularx}
\label{tab:verifier_datasets}
\end{table*}

\subsection{Course Planning}


In Table \ref{tab:solver_datasets}, LLMs perform well in generating complete plans with all sections included when the problem size is small. However, as the problem description becomes longer, they struggle to identify the total number of sections, often resulting in incomplete plans. Across all task difficulties, LLMs fail to produce feasible plans, let alone optimal ones, even if the plans are complete. Among these models, Claude-3.5-Sonnet demonstrates the best overall performance in understanding natural language instructions and creating complete plans.


Table \ref{tab:verifier_datasets} summarizes the performance of LLMs as verifiers in course planning tasks across varying difficulty levels. Overall, the results show that all models struggle significantly to evaluate both the feasibility and optimality of plans, with performance dropping as task difficulty increases.

GPT-4o outperforms other models in detecting feasible plans, achieving the highest rate of 61.25\% in easy level. However, the ability to identify optimal solution is not as expected. Same situation happens to claude and DeepSeek where the ability to detect feasible plans is much better than to detect optimal plans while DeepSeek performs the worst among them. For all models, when problems become harder, the ability to verify plans is still limited, remaining challenge for further exploration.

From Table \ref{tab:heuristic_comparison_difficulty}, it is evident that Claude consistently outperforms other models across all evaluation settings. As task difficulty increases, the performance of all models deteriorates. However, while Claude experiences a relatively mild decline, GPT-4o and DeepSeek-Chat exhibit steeper drops in performance, highlighting their greater sensitivity to task complexity. In contrast, the fine-tuned reward model LoRA underperforms across all configurations, further emphasizing the challenges faced by simpler or more narrowly tuned models in handling heuristic-based evaluations.

The comparison between zero-shot and one-shot evaluations underscores the benefits of applying comparative heuristic functions. When provided with a heuristic, language models are better guided in their reasoning, reducing the likelihood of errors by narrowing their focus in alignment with the heuristic's direction. Furthermore, the comparison between the one-shot and reward model configurations demonstrates the effectiveness of heuristic guidance, suggesting that incorporating such methods could serve as a promising direction for future research and development.


\begin{table*}[t]
\centering
\caption{Heuristic Performance Comparison across Multiple Datasets}
\renewcommand{\arraystretch}{1.3}
\label{tab:heuristic_4candidate}
\begin{tabular}{p{3.5cm} l cccc}
\hline
\textbf{Dataset Name} & \textbf{Model} & \textbf{Hit@1} & \textbf{Hit@2} & \textbf{Hit@3} & \textbf{Comparison Accuracy} \\
\hline
\multirow{3}{*}{Course Planning-Easy} 
    & DeepSeek & 0.2725 & 0.5650 & 0.7800 & 0.4575 \\ 
    & GPT-4o   & 0.2800 & 0.5800 & 0.8000 & 0.5125 \\ 
    & Claude   & 0.4075 & 0.7125 & 0.9225 & 0.7525 \\ 
\hline
\multirow{3}{*}{Course Planning-Medium} 
    & DeepSeek & 0.2125 & 0.5200 & 0.7750 & 0.4650 \\ 
    & GPT-4o   & 0.2125 & 0.5025 & 0.7400 & 0.4725 \\ 
    & Claude   & 0.3400 & 0.6775 & 0.8925 & 0.7375 \\ 
\hline
\multirow{3}{*}{Course Planning-Hard} 
    & DeepSeek & 0.2400 & 0.4875 & 0.7575 & 0.4175 \\ 
    & GPT-4o   & 0.2425 & 0.5125 & 0.7500 & 0.5025 \\ 
    & Claude   & 0.4050 & 0.6850 & 0.9150 & 0.7250 \\ 
\hline
\multirow{2}{*}{Fitness Planning} 
    & DeepSeek & 0.5100 & 0.8100 & 0.9300 & 0.7000 \\ 
    & GPT-4o   & 0.5000 & 0.7900 & 0.9200 & 0.7100 \\ 
\hline
\multirow{2}{*}{Travel Planning} 
    & DeepSeek & 0.4800 & 0.7556 & 0.9200 & 0.5846 \\ 
    & GPT-4o   & 0.5911 & 0.8400 & 0.9556 & 0.5538 \\ 
\hline
\end{tabular}
\end{table*}
\subsection{Travel Planning}

In Table \ref{tab: simple_solver}, GPT-4o demonstrates a significant performance advantage over DeepSeek across both evaluation settings. However, despite this relative superiority, neither GPT-4o nor DeepSeek achieves results that can be deemed satisfactory or robust for practical applications. When the Chain-of-Thought (CoT) strategy is applied, GPT-4o exhibits a modest improvement in performance, leveraging its step-by-step reasoning capability to better address the problem space. In contrast, DeepSeek's performance deteriorates further, failing to produce any correct plans under this strategy. Both models struggle with detecting implicit constraints that are intuitive and readily understood by humans, and they consistently fail to adhere to all specified constraints, underscoring the limitations of current methodologies in handling complex, constraint-rich planning tasks.

In Table \ref{tab:verifier_datasets}, the performance of LLMs as verifiers on the Travel Planning dataset highlights significant limitations. GPT-4o achieves a pass rate of 0.4550, slightly underperforming compared to DeepSeek, which records a pass rate of 0.4890. Despite these marginal differences, neither model demonstrates acceptable verification capabilities. The observed results fall short of expectations for practical applications, as the dataset is balanced, and the models' performance is barely above a random baseline.

The suboptimal outcomes suggest fundamental challenges in the models' ability to recognize and validate the complex constraints inherent in travel planning tasks. This underscores the limitations of current LLM-based approaches in systematically evaluating and ensuring the feasibility and optimality of solutions in constraint-heavy environments. Further exploration is needed to enhance the models' reasoning and constraint detection capabilities to achieve satisfactory results in such tasks.

In Table \ref{tab:heuristic_4candidate}, GPT-4o demonstrates overall superior performance compared to DeepSeek across the Hit@1, Hit@2, and Hit@3 metrics, underscoring its enhanced capability to pinpoint and prioritize the most optimal solutions within the candidate set. However, GPT-4o lags behind DeepSeek in comparison accuracy, which highlights DeepSeek's greater proficiency in recognizing the second-best plan and distinguishing fine-grained differences among candidates. This is further supported by the narrower gap between Hit@2 (0.8400 - 0.7556) and Hit@1 (0.5911 - 0.4800) for DeepSeek, showcasing its consistency in evaluating suboptimal solutions effectively.

\section{Discussion}
This study investigated the performance of LLMs on planning tasks across three distinct domains: travel planning, course planning, and fitness planning. A common thread across these domains is the crucial role of constraint following in generating successful solutions. This discussion will delve into the observed limitations of LLMs in adhering to constraints and explore the surprising challenges encountered in verifying the generated plans.

\subsection{Limitations on Constraint Following}

Constraint satisfaction is paramount for effective problem-solving in the chosen benchmarks. Travel planning often necessitates the understanding of implicit (commonsense) constraints, where requirements are not explicitly stated but should be inferred. Course planning, on the other hand, presents explicit constraints, such as non-overlapping time slots and classroom capacity limits. Lastly, fitness planning involves both explicit hard constraints (e.g., maximum exercise duration) and implicit user preferences.

Our experiments revealed that LLMs, despite their impressive capabilities, can struggle to adhere to these constraints. We identified three primary contributing factors:

\begin{enumerate}[leftmargin=*]
\item Lack of Constraint Awareness: LLMs may fail to recognize the existence of a constraint, particularly in cases of implicit constraints. This was evident in the travel planner dataset, where agents occasionally planned multiple visits to the same attraction without being explicitly instructed to do so. While these constraints were not explicitly provided, they represent fundamental commonsense knowledge expected of a competent planning agent. Future work could explore a multi-step approach where agents first explicitly outline implicit requirements before generating a plan.
\item Constraint Neglect: Even when aware of constraints, LLMs might disregard them during plan generation. This was observed in the fitness planning dataset, where models like Deepseek occasionally ignored the explicitly stated maximum exercise time. This suggests a need for more controlled generative processes or enhanced agentic frameworks that explicitly reinforce constraint adherence during plan generation.
\item Difficulty with Global Constraints: Certain global constraints are inherently difficult to follow. Course planning exemplifies this with its global constraint satisfaction problem, where assigning a class section to a time slot can impact all other sections. The autoregressive nature of GPT models makes it inherently difficult to solve this category of problems. By design, it is very difficult for an autoregressive model to attend to future states (though not impossible). This points to the potential need for methodologies that extend beyond autoregressive modeling, such as incorporating lookahead mechanisms or exploring alternative architectures.
\end{enumerate}

\subsection{The Unexpected Difficulty of Verification}

Algorithmically, many problems, including those investigated in this study, are easier to verify than to solve. However, our findings revealed a surprising difficulty for LLMs in verifying solutions. This was particularly evident in the course planning task, where LLMs struggled to determine whether a generated plan adhered to time and capacity constraints. The challenge was even more pronounced in the travel planner dataset, where models failed to reliably evaluate solution correctness and performed poorly even when tasked with ranking two solutions.

The ability to automatically verify solutions is essential for real-world applications. While tasks like course planning may allow for straightforward implementation of verifiers, many real-life scenarios, such as travel planning, do not. When evaluation criteria are subjective and verification relies on conceptual understanding rather than purely mathematical checks (e.g., assessing adherence to commonsense versus verifying time conflict absence), verification itself becomes a formidable challenge. This highlights the need for research into robust and adaptable verification methods that can handle the nuances of real-world planning problems.

\section{Conclusion}
This study offers a comprehensive evaluation of the role of large language models (LLMs) in addressing planning problems. We specifically investigate the effectiveness of LLMs as planners, solution verifiers, and sources of heuristic feedback for refining intermediate solutions. Our findings indicate that while LLMs struggle to directly produce accurate plans, they excel at generating comparative feedback that can act as a heuristic function. This evaluation framework offers valuable guidance for the development of future LLM-driven tree-search algorithms for a variety of planning and reasoning tasks. Furthermore, we introduce a new benchmark for assessing LLMs' capacity to adapt to user preferences in real-time, a crucial capability with broad practical implications.

\newpage
\bibliographystyle{unsrtnat}
\bibliography{references}

\begin{thebibliography}{19}
\providecommand{\natexlab}[1]{#1}
\providecommand{\url}[1]{\texttt{#1}}
\expandafter\ifx\csname urlstyle\endcsname\relax
  \providecommand{\doi}[1]{doi: #1}\else
  \providecommand{\doi}{doi: \begingroup \urlstyle{rm}\Url}\fi

\bibitem[Snell et~al.(2024)Snell, Lee, Xu, and Kumar]{snell2024scalingllmtesttimecompute}
Charlie Snell, Jaehoon Lee, Kelvin Xu, and Aviral Kumar.
\newblock Scaling llm test-time compute optimally can be more effective than scaling model parameters, 2024.
\newblock URL \url{https://arxiv.org/abs/2408.03314}.

\bibitem[Hao et~al.(2023)Hao, Gu, Ma, Hong, Wang, Wang, and Hu]{hao2023reasoninglanguagemodelplanning}
Shibo Hao, Yi~Gu, Haodi Ma, Joshua~Jiahua Hong, Zhen Wang, Daisy~Zhe Wang, and Zhiting Hu.
\newblock Reasoning with language model is planning with world model, 2023.
\newblock URL \url{https://arxiv.org/abs/2305.14992}.

\bibitem[Yao et~al.(2023{\natexlab{a}})Yao, Yu, Zhao, Shafran, Griffiths, Cao, and Narasimhan]{yao2023treethoughtsdeliberateproblem}
Shunyu Yao, Dian Yu, Jeffrey Zhao, Izhak Shafran, Thomas~L. Griffiths, Yuan Cao, and Karthik Narasimhan.
\newblock Tree of thoughts: Deliberate problem solving with large language models, 2023{\natexlab{a}}.
\newblock URL \url{https://arxiv.org/abs/2305.10601}.

\bibitem[Long(2023)]{long2023largelanguagemodelguided}
Jieyi Long.
\newblock Large language model guided tree-of-thought, 2023.
\newblock URL \url{https://arxiv.org/abs/2305.08291}.

\bibitem[Shinn et~al.(2023)Shinn, Cassano, Berman, Gopinath, Narasimhan, and Yao]{shinn2023reflexion}
Noah Shinn, Federico Cassano, Edward Berman, Ashwin Gopinath, Karthik Narasimhan, and Shunyu Yao.
\newblock Reflexion: Language agents with verbal reinforcement learning.
\newblock In \emph{Proceedings of the 37th Conference on Neural Information Processing Systems (NeurIPS 2023)}, 2023.

\bibitem[Yao et~al.(2023{\natexlab{b}})Yao, Zhao, Yu, Du, Shafran, Narasimhan, and Cao]{yao2023react}
Shunyu Yao, Jeffrey Zhao, Dian Yu, Nan Du, Izhak Shafran, Karthik Narasimhan, and Yuan Cao.
\newblock React: Synergizing reasoning and acting in language models.
\newblock In \emph{Proceedings of the International Conference on Learning Representations (ICLR 2023)}, 2023{\natexlab{b}}.

\bibitem[Huang et~al.(2022)Huang, Abbeel, Pathak, and Mordatch]{huang2022language}
Wenlong Huang, Pieter Abbeel, Deepak Pathak, and Igor Mordatch.
\newblock Language models as zero-shot planners: Extracting actionable knowledge for embodied agents.
\newblock In \emph{International Conference on Machine Learning}, pages 9118--9147. PMLR, 2022.

\bibitem[Song et~al.(2023)Song, Wu, Washington, Sadler, Chao, and Su]{song2023llmplanner}
Chan~Hee Song, Jiaman Wu, Clayton Washington, Brian~M. Sadler, Wei-Lun Chao, and Yu~Su.
\newblock Llm-planner: Few-shot grounded planning for embodied agents with large language models.
\newblock In \emph{Proceedings of the IEEE International Conference on Computer Vision (ICCV)}, 2023.

\bibitem[Yang et~al.(2024)Yang, Chen, Zhang, Yuan, Chen, Richardson, Xiao, and Yang]{yang2024selfgoallanguageagentsknow}
Ruihan Yang, Jiangjie Chen, Yikai Zhang, Siyu Yuan, Aili Chen, Kyle Richardson, Yanghua Xiao, and Deqing Yang.
\newblock Selfgoal: Your language agents already know how to achieve high-level goals, 2024.
\newblock URL \url{https://arxiv.org/abs/2406.04784}.

\bibitem[Hazra et~al.(2024)Hazra, Zuidberg Dos~Martires, and De~Raedt]{hazra2024saycanpay}
Rishi Hazra, Pedro Zuidberg Dos~Martires, and Luc De~Raedt.
\newblock Saycanpay: Heuristic planning with large language models using learnable domain knowledge.
\newblock \emph{arXiv preprint arXiv:2308.12682}, 2024.

\bibitem[Singh et~al.(2022)Singh, Blukis, Mousavian, Goyal, Xu, Tremblay, Fox, Thomason, and Garg]{singh2022progprompt}
Ishika Singh, Valts Blukis, Arsalan Mousavian, Ankit Goyal, Danfei Xu, Jonathan Tremblay, Dieter Fox, Jesse Thomason, and Animesh Garg.
\newblock Progprompt: Generating situated robot task plans using large language models.
\newblock In \emph{Proceedings of the 36th Conference on Neural Information Processing Systems (NeurIPS)}, 2022.

\bibitem[Sun et~al.(2023)Sun, Zhuang, Kong, Dai, and Zhang]{sun2023adaplanneradaptiveplanningfeedback}
Haotian Sun, Yuchen Zhuang, Lingkai Kong, Bo~Dai, and Chao Zhang.
\newblock Adaplanner: Adaptive planning from feedback with language models, 2023.
\newblock URL \url{https://arxiv.org/abs/2305.16653}.

\bibitem[Li et~al.(2022)Li, Puig, Paxton, Du, Wang, Fan, Chen, Huang, Aky{"u}rek, Anandkumar, Andreas, Mordatch, Torralba, and Zhu]{li2022lid}
Shuang Li, Xavier Puig, Chris Paxton, Yilun Du, Clinton Wang, Linxi Fan, Tao Chen, De-An Huang, Ekin Aky{"u}rek, Anima Anandkumar, Jacob Andreas, Igor Mordatch, Antonio Torralba, and Yuke Zhu.
\newblock Pre-trained language models for interactive decision-making.
\newblock In \emph{36th Conference on Neural Information Processing Systems (NeurIPS 2022)}, 2022.

\bibitem[Wang et~al.(2023)Wang, Cai, Chen, Liu, Ma, and Liang]{wang2023describe}
Zihao Wang, Shaofei Cai, Guanzhou Chen, Anji Liu, Xiaojian Ma, and Yitao Liang.
\newblock Describe, explain, plan and select: Interactive planning with large language models.
\newblock In \emph{37th Conference on Neural Information Processing Systems (NeurIPS)}, 2023.
\newblock URL \url{https://github.com/CraftJarvis/MC-Planner}.

\bibitem[OpenAI(2024)]{openai2024learning}
OpenAI.
\newblock Learning to reason with llms, 2024.
\newblock URL \url{https://openai.com/index/learning-to-reason-with-llms/}.
\newblock Accessed: 2024-11-27.

\bibitem[Brown et~al.(2024)Brown, Juravsky, Ehrlich, Clark, Le, Ré, and Mirhoseini]{brown2024largelanguagemonkeysscaling}
Bradley Brown, Jordan Juravsky, Ryan Ehrlich, Ronald Clark, Quoc~V. Le, Christopher Ré, and Azalia Mirhoseini.
\newblock Large language monkeys: Scaling inference compute with repeated sampling, 2024.
\newblock URL \url{https://arxiv.org/abs/2407.21787}.

\bibitem[Xie et~al.(2024)Xie, Zhang, Chen, Zhu, Lou, Tian, Xiao, and Su]{xie2024travelplanner}
Jian Xie, Kai Zhang, Jiangjie Chen, Tinghui Zhu, Renze Lou, Yuandong Tian, Yanghua Xiao, and Yu~Su.
\newblock Travelplanner: A benchmark for real-world planning with language agents.
\newblock In \emph{Forty-first International Conference on Machine Learning}, 2024.

\bibitem[Shin et~al.(2023)Shin, Hsieh, and Kim]{shin2023planfittingtailoringpersonalizedexercise}
Donghoon Shin, Gary Hsieh, and Young-Ho Kim.
\newblock Planfitting: Tailoring personalized exercise plans with large language models, 2023.
\newblock URL \url{https://arxiv.org/abs/2309.12555}.

\bibitem[Kasneci et~al.(2023)Kasneci, Se{\ss}ler, K{\"u}chemann, Bannert, Dementieva, Fischer, Gasser, Groh, G{\"u}nnemann, H{\"u}llermeier, et~al.]{kasneci2023chatgpt}
Enkelejda Kasneci, Kathrin Se{\ss}ler, Stefan K{\"u}chemann, Maria Bannert, Daryna Dementieva, Frank Fischer, Urs Gasser, Georg Groh, Stephan G{\"u}nnemann, Eyke H{\"u}llermeier, et~al.
\newblock Chatgpt for good? on opportunities and challenges of large language models for education.
\newblock \emph{Learning and individual differences}, 103:\penalty0 102274, 2023.

\end{thebibliography}

\newpage
\appendix

\onecolumn
\section{Result for LLM Using Heuristic Function in Course Planning}
\begin{table*}[h!]
\centering
\renewcommand{\arraystretch}{1.2}

\begin{tabular}{llccccc}
\hline
\textbf{Difficulty} & \textbf{Candidates}    & \textbf{Method}        & \textbf{Model}          & \textbf{Hit@1} & \textbf{Hit@2} & \textbf{Hit@3} \\
\hline
\multirow{14}{*}{Easy} 
    & \multirow{7}{*}{2 Candidates} 
        & \multirow{3}{*}{Zero-shot} 
            & GPT-4o                  & 0.5125              & -              & -             \\
        &                        &                        & Claude-3.5-Sonnet       & 0.7525              & -              & -             \\
        &                        &                        & DeepSeek-Chat           & 0.4575              & -              & -             \\
        \cline{3-7}
        &                        & \multirow{3}{*}{One-shot} 
            & GPT-4o                  & 0.7975              & -              & -             \\
        &                        &                        & Claude-3.5-Sonnet       & 0.8275              & -              & -             \\
        &                        &                        & DeepSeek-Chat           & 0.730               & -              & -             \\
        \cline{3-7}
        &                        & Reward Model          & LoRA                   & 0.2825              & -              & -             \\
    \cline{2-7}
    & \multirow{7}{*}{4 Candidates} 
        & \multirow{3}{*}{Zero-shot} 
            & GPT-4o                  & 0.28               & 0.58               & 0.8             \\
        &                        &                        & Claude-3.5-Sonnet       & 0.4075             & 0.7125            & 0.9225             \\
        &                        &                        & DeepSeek-Chat           & 0.2725             & 0.565             & 0.78             \\
        \cline{3-7}
        &                        & \multirow{3}{*}{One-shot} 
            & GPT-4o                  & 0.45               & 0.7775             & 0.93             \\
        &                        &                        & Claude-3.5-Sonnet       & 0.4775             & 0.745             & 0.935             \\
        &                        &                        & DeepSeek-Chat           & 0.3875             & 0.6975            & 0.9075             \\
        \cline{3-7}
        &                        & Reward Model          & LoRA                   & 0.1725             & 0.4225            & 0.7075             \\
\hline
\multirow{14}{*}{Medium} 
    & \multirow{7}{*}{2 Candidates} 
        & \multirow{3}{*}{Zero-shot} 
            & GPT-4o                  & 0.4725              & -              & -             \\
        &                        &                        & Claude-3.5-Sonnet       & 0.7375              & -              & -             \\
        &                        &                        & DeepSeek-Chat           & 0.465              & -              & -             \\
        \cline{3-7}
        &                        & \multirow{3}{*}{One-shot} 
            & GPT-4o                  & 0.7825              & -              & -             \\
        &                        &                        & Claude-3.5-Sonnet       & 0.7975              & -              & -             \\
        &                        &                        & DeepSeek-Chat           & 0.6375               & -              & -             \\
        \cline{3-7}
        &                        & Reward Model          & LoRA                   & 0.245              & -              & -             \\
    \cline{2-7}
    & \multirow{7}{*}{4 Candidates} 
        & \multirow{3}{*}{Zero-shot} 
            & GPT-4o                  & 0.2125               & 0.5025               & 0.74             \\
        &                        &                        & Claude-3.5-Sonnet       & 0.34             & 0.6775            & 0.8925             \\
        &                        &                        & DeepSeek-Chat           & 0.2125             & 0.52             & 0.775             \\
        \cline{3-7}
        &                        & \multirow{3}{*}{One-shot} 
            & GPT-4o                  & 0.3775               & 0.7025             & 0.895             \\
        &                        &                        & Claude-3.5-Sonnet       & 0.4675             & 0.745             & 0.9325             \\
        &                        &                        & DeepSeek-Chat           & 0.3575             & 0.6675            & 0.89             \\
        \cline{3-7}
        &                        & Reward Model          & LoRA                   & 0.2125             & 0.44            & 0.7425             \\
\hline
\multirow{14}{*}{Hard} 
    & \multirow{7}{*}{2 Candidates} 
        & \multirow{3}{*}{Zero-shot} 
            & GPT-4o                  & 0.5025              & -              & -             \\
        &                        &                        & Claude-3.5-Sonnet       & 0.725              & -              & -             \\
        &                        &                        & DeepSeek-Chat           & 0.4175              & -              & -             \\
        \cline{3-7}
        &                        & \multirow{3}{*}{One-shot} 
            & GPT-4o                  & 0.70              & -              & -             \\
        &                        &                        & Claude-3.5-Sonnet       & 0.7675              & -              & -             \\
        &                        &                        & DeepSeek-Chat           & 0.675               & -              & -             \\
        \cline{3-7}
        &                        & Reward Model          & LoRA                   & 0.2425              & -              & -             \\
    \cline{2-7}
    & \multirow{7}{*}{4 Candidates} 
        & \multirow{3}{*}{Zero-shot} 
            & GPT-4o                  & 0.2425               & 0.5125               & 0.75             \\
        &                        &                        & Claude-3.5-Sonnet       & 0.405             & 0.685            & 0.915             \\
        &                        &                        & DeepSeek-Chat           & 0.24             & 0.4875             & 0.7575             \\
        \cline{3-7}
        &                        & \multirow{3}{*}{One-shot} 
            & GPT-4o                  & 0.4375               & 0.7175             & 0.885             \\
        &                        &                        & Claude-3.5-Sonnet       & 0.44             & 0.7325             & 0.935             \\
        &                        &                        & DeepSeek-Chat           & 0.2275             & 0.5275            & 0.775             \\
        \cline{3-7}
        &                        & Reward Model          & LoRA                   & 0.21             & 0.46            & 0.7325             \\
\hline
\end{tabular}
\vspace{0.5em}
\caption{Heuristic Performance Comparison Across Difficulty Levels and Evaluation Settings}
\label{tab:heuristic_comparison_difficulty}
\end{table*}
\twocolumn

\onecolumn
\section{Detailed results for LLM as Solver}
\begin{table*}[h!]
\centering
\caption{LLMs Work as Solver Across Multiple Datasets}
\renewcommand{\arraystretch}{1.3} 
\begin{tabularx}{\textwidth}{@{}llXccccccc@{}}
\toprule
\textbf{Dataset} & \textbf{Task Level} & \textbf{Condition} & \textbf{Iterations} & \textbf{Model} & \textbf{Comp.} & \textbf{Feas.} & \textbf{Opt.} & \textbf{Diversity} & \textbf{PassRate} \\ 
\midrule
\multirow{9}{*}{\makecell{Course\\ Planning}} 
    & \multirow{3}{*}{Easy} 
        & - & - & GPT-4o        & 1.000 & 0.09   & 0.2778 & -     & -       \\ 
    &                   & - & - & Claude        & 1.000 & 0.1025 & 0.7073 & -     & -       \\ 
    &                   & - & - & DeepSeek      & 0.9975 & 0.0576 & 0.2174 & -     & -       \\ 
\cmidrule(lr){2-10}
    & \multirow{3}{*}{Medium} 
        & - & - & GPT-4o        & 0.985 & 0.0228 & 0.3334 & -     & -       \\ 
    &                   & - & - & Claude        & 0.985 & 0.0305 & 0.9167 & -     & -       \\ 
    &                   & - & - & DeepSeek      & 0.9575 & 0.0052 & 0.5000    & -     & -       \\ 
\cmidrule(lr){2-10}
    & \multirow{3}{*}{Hard} 
        & - & - & GPT-4o        & 0.6550 & 0.000   & 0.0000   & -     & -       \\ 
    &                   & - & - & Claude        & 0.6550 & 0.0038 & 1.0000   & -     & -       \\ 
    &                   & - & - & DeepSeek      & 0.6275 & 0.0040 & 1.0000   & -     & -       \\ 
\midrule
\multirow{13}{*}{\makecell{Fitness\\ Planning}} 
    & \multirow{13}{*}{-} 
        & \multirow{6}{*}{\makecell[c]{Without \\Dynamic\\ Constraint}}
        & 5  & GPT-4o & - & 0.7370 & 0.7660 & 0.2330 & - \\ 
    &     &  & 5  & DeepSeek & - & 0.0000   & 0.0000   & 0.0000   & - \\ 
\cmidrule(lr){4-10}
    &     &  & 10 & GPT-4o & - & 0.6670 & 0.8490 & 0.3330 & - \\ 
    &     &  & 10 & DeepSeek & - & 0.0000   & 0.0000   & 0.0000   & - \\ 
\cmidrule(lr){4-10}
    &     &  & 20 & GPT-4o & - & 0.7500 & 0.8520 & 0.3370 & - \\ 
    &     &  & 20 & DeepSeek & - & 0.0000   & 0.0000   & 0.0000   & - \\ 
\cmidrule(lr){3-10}
    &     & \multirow{6}{*}{\makecell{With \\Dynamic\\ Constraint}} 
        & 5  & GPT-4o & - & 0.4130 & 0.5000 & 0.4670 & - \\ 
    &     &  & 5  & DeepSeek & - & 0.0000   & 0.0000   & 0.0000   & - \\ 
\cmidrule(lr){4-10}
    &     &  & 10 & GPT-4o & - & 0.4330 & 0.6140 & 0.4830 & - \\ 
    &     &  & 10 & DeepSeek & - & 0.0000   & 0.0000   & 0.0000   & - \\ 
\cmidrule(lr){4-10}
    &     &  & 20 & GPT-4o & - & 0.6300 & 0.7050 & 0.5470 & - \\ 
    &     &  & 20 & DeepSeek & - & 0.0000   & 0.0000   & 0.0000   & - \\ 
\midrule
\multirow{2}{*}{\makecell{Travel\\ Planning}} 
    & \multirow{2}{*}{-} 
        & Direct & - & GPT-4o        & -     & -      & -      & -     & 0.0720   \\ 
    &                   & Direct & - & DeepSeek      & -     & -      & -      & -     & 0.0050   \\ 
\cmidrule(lr){3-10}
    & \multirow{2}{*}{-} 
        & CoT & - & GPT-4o        & -     & -      & -      & -     & 0.0830   \\ 
    &                   & CoT & - & DeepSeek      & -     & -      & -      & -     & 0.0000   \\ 
\bottomrule
\end{tabularx}
\label{tab:solver_datasets}
\end{table*}
\twocolumn
\newpage

\section{Fitness Planning Example}
The user, Joe, has the following constraints:

\textbf{Gym Access}: No\\
\textbf{Available Time}: 60 minutes\\
\textbf{Fitness Goal}: Lose weight\\
\textbf{Stamina Level}: Medium\\
\textbf{Exercise Preferences}: Joe prefers aerobic exercises over anaerobic ones, with a preference score of 5 for Jogging, Jump Rope, Push-Up, etc.

The agent selects from a set of predefined exercises. Table \ref{tab:exercises} shows a few sample exercises from the available options, with Time recorded in minutes and Int. representing Intensity measured in High (H), Medium (M), and Low (L).

\begin{table}[h!]
\centering
\begin{tabular}{@{}lllll@{}}
\toprule
\textbf{Exercise}     & \textbf{Time} & \textbf{Int.} & \textbf{Gym} & \textbf{Category} \\ \midrule
Jogging               & 30                  & L                & No                  & Aerobic           \\
Cycling               & 45                  & M             & Yes                 & Aerobic           \\
Swimming              & 60                  & H               & Yes                 & Aerobic           \\
Jump Rope             & 15                  & H               & No                  & Aerobic           \\
Push-Up               & 2                   & M             & No                  & Anaerobic         \\
Bench Press           & 5                   & H               & Yes                 & Anaerobic         \\
Shoulder Shrugs       & 5                   & L                & No                  & Anaerobic         \\
Lunges                & 5                   & M             & No                  & Anaerobic         \\ \bottomrule
\end{tabular}
\caption{Sample of Exercises Available}
\label{tab:exercises}
\end{table}

\textbf{Initial Plan and Feedback}: The agent generates an initial workout plan, as shown in Table \ref{tab:plan1}. The plan includes the number of repetitions for each exercise. Joe provides feedback after executing the plan.

\begin{table}[h]
\centering
\begin{tabular}{@{}ll@{}}
\toprule
\textbf{Exercise}  & \textbf{Reps} \\ \midrule
Jogging            & 1             \\
Jump Rope          & 2             \\
Push-Up            & 2             \\
Shoulder Shrugs    & 1             \\
Lunges             & 2             \\ \bottomrule
\end{tabular}
\caption{Initial Workout Plan}
\label{tab:plan1}
\end{table}

\textbf{User Feedback}: Joe reported that the total time required was 79 minutes, which exceeded his available time of 60 minutes. Therefore, this plan is \textit{inadmissible}. (Note that the \textbf{User Feedback} will be appended to the message list.)

\textbf{Refined Plan and Feedback}: Based on Joe's feedback, the agent refines the plan harnessing its reasoning ability to fit within Joe's time constraints, as shown in Table \ref{tab:plan2}. (The \textbf{reasoning} will also be appended to the message list.)

\textbf{User Feedback}: Joe found this plan to be executable within his available time and provided a satisfaction score of 8.33 out of 10. (And the iteration goes on...)

\begin{table}[b]
\centering
\begin{tabular}{@{}ll@{}}
\toprule
\textbf{Exercise}  & \textbf{Reps} \\ \midrule
Jogging            & 1             \\
Jump Rope          & 1             \\
Push-Up            & 3             \\
Lunges             & 1             \\ \bottomrule
\end{tabular}
\caption{Refined Workout Plan}
\label{tab:plan2}
\end{table}

\onecolumn
\section{Course Planning Example}
\lstset{
  basicstyle=\ttfamily\scriptsize, 
  breaklines=true,                  
  breakindent=0pt,                  
  frame=single,                     
  xleftmargin=2pt, xrightmargin=2pt 
}
\begin{lstlisting}[]
{
    "raw_problem": {
        "Class Periods": {
            "Course 1": {
                "Section 1": "['Monday', 'Thursday'] at 11:30AM-12:45PM",
                "Section 2": "['Tuesday', 'Monday'] at 2:30PM-3:45PM",
                "Section 3": "['Monday', 'Thursday'] at 8:30AM-9:45AM"
            },
            "Course 2": {
                "Section 1": "['Tuesday', 'Thursday'] at 11:30AM-12:45PM",
                "Section 2": "['Tuesday', 'Friday'] at 5:30PM-6:45PM",
                "Section 3": "['Friday', 'Tuesday'] at 4:00PM-5:15PM"
            },
            "Course 3": {
                "Section 1": "['Tuesday', 'Monday'] at 5:30PM-6:45PM",
                "Section 2": "['Monday', 'Friday'] at 11:30AM-12:45PM"
            },
        },
        "number_of_seats": {
            "Course 1": {
                "Section 1": 21,
                "Section 2": 21,
                "Section 3": 30
            },
            "Course 2": {
                "Section 1": 29,
                "Section 2": 25,
                "Section 3": 24
            },
        },
        "Classrooms": {
            "classroom 1": 28,
            "classroom 2": 32
        }
    },
    "text_description": "The course schedule is organized as follows: ... (omitted for brevity)",
    "solution": {
        "Course 1": {
            "Section 1": {
                "room": "classroom 1",
                "seat_diff": 7
            },
            "Section 2": {
                "room": "classroom 1",
                "seat_diff": 7
            },
            "Section 3": {
                "room": "classroom 2",
                "seat_diff": 2
            }
        },
        "Course 2": {
            "Section 1": {
                "room": "classroom 2",
                "seat_diff": 3
            },
            "Section 2": {
                "room": "classroom 1",
                "seat_diff": 3
            },
            "Section 3": {
                "room": "classroom 1",
                "seat_diff": 4
            }
        },
        "Course 3": {
            "Section 1": {
                "room": "classroom 2",
                "seat_diff": 3
            },
            "Section 2": {
                "room": "classroom 2",
                "seat_diff": 3
            }
        },
    },
    "optimal_score": 32.0
}
\end{lstlisting}
\twocolumn

\end{document}